\documentclass[a4paper,twoside]{article}

\usepackage[T1]{fontenc}
\usepackage{graphicx}
\usepackage{epsfig}
\usepackage{subcaption}
\usepackage{calc}

\usepackage{amsmath}
\usepackage{amssymb}
\usepackage{amsfonts}
\usepackage{amstext}
\usepackage{amsthm}

\usepackage{xcolor}
\usepackage{booktabs}
\usepackage{tabularx}
\usepackage{tikz}

\usepackage{enumitem}
\usepackage{comment}
\usepackage{caption}
\usepackage{placeins}
\usepackage{multicol}

\usepackage{pslatex}
\usepackage{apalike}

\usepackage{algorithm}
\usepackage{algpseudocode}

\usepackage[algo2e]{algorithm2e}

\usepackage[bottom]{footmisc}

\usepackage{hyperref}
\hypersetup{
    colorlinks=true,
    linkcolor=black,
    citecolor=black,
    urlcolor=black
}

\usepackage[nameinlink,capitalize,noabbrev]{cleveref}
\usepackage{fontawesome5}

\usepackage{SCITEPRESS}

\newcommand{\cmark}{\checkmark}
\newcommand{\xmark}{\ensuremath{\times}}

\crefformat{equation}{\textcolor{blue}{#2#1#3}}

\begin{document}

\title{Uncertainty-Aware Candidate Generation for Entity Linking in Cyber Threat Intelligence}


\renewcommand{\orcidAuthor}[1]{}
\author{
\authorname{
Nasser Alkhulaifi\sup{1,2}\orcidAuthor{0000-0002-6146-5114},
Direnc Pekaslan\sup{1}\orcidAuthor{0000-0003-2035-4036} and
Christian Wagner\sup{1}\orcidAuthor{0000-0002-6121-9722}
}
\affiliation{
\sup{1}LUCID Lab, School of Computer Science,
University of Nottingham, Nottingham, UK
}
\affiliation{
\sup{2}COL Lab,
School of Computer Science,
University of Nottingham, UK
}
\email{
\{Nasser.Alkhulaifi,Direnc.Pekaslan1,Christian.Wagner\}@nottingham.ac.uk
}
}



\abstract{
Entity Resolution (ER) pipelines often collapse ambiguity into early hard decisions: upstream Named Entity Recognition (NER) selects a single span and type, and downstream Entity Linking (EL) treats that choice as ground truth. This discards valuable information about the ambiguity of the underlying data (e.g., span boundary disagreement and type confusion), leading to brittle systems that fail unpredictably when faced with uncertainty in inputs. Therefore, this work presents a proof-of-concept for uncertainty-aware ER that treats uncertainty as native data that should be preserved and exploited throughout ER pipelines. We (i) capture uncertainty signals in the NER stage and (ii) leverage these signals into the EL stage to improve candidate retrieval under ambiguity. Using a BERT-based NER model augmented with two auxiliary span-boundary prediction heads, we apply Monte Carlo (MC) dropout at inference time to derive three mention-level proxy signals: the MC-averaged start-boundary probability, the MC-averaged end-boundary probability, and De Luca--Termini fuzzy entropy computed over the Begin--Inside--Outside (BIO) predictive distribution. These signals are used to adapt the fuzzy cutoff in candidate generation (CG) for EL. Because CG is treated as a blocking step, we evaluate retrieval recall. On the AnnoCTR dataset linked to the MITRE ATT\&CK knowledge base, the boundary-aware and dual-signal fuzzy linking approaches improve Recall@10 from 0.6186 to 0.7966. The contribution lies not in an end-to-end optimised ER pipeline but in establishing uncertainty preservation as a viable and beneficial component of ER pipelines, enabling downstream EL to make better-informed decisions under uncertainty.
}


\onecolumn
\maketitle
\normalsize
\setcounter{footnote}{0}
\vfill


\section{Introduction and Background}
\label{Introduction and Background}

Entity Resolution (ER) aims to determine when different textual mentions or records refer to the same real-world entity and to align them to a canonical representation, typically within or across Knowledge Bases (KBs) \cite{Zhao2019_Auto_EM,Christophides2020_end_to_end_ER}. Accurate ER is a prerequisite for the construction and maintenance of high-quality Knowledge Graphs (KGs), since unresolved duplicates and incorrect merges propagate errors to downstream analytics, search, and reasoning \cite{Wang2021_Knowledge_graph_quality,Zhong2023_Automatic_Knowledge_Graph}. ER for unstructured text is commonly operationalised as a pipeline: entities are first detected and typed via Named Entity Recognition (NER), then grounded (linked) to a KB via Entity Linking (EL), enabling richer structured context for subsequent retrieval and inference \cite{Christophides2020_end_to_end_ER}.


The upstream task of ER is NER, a natural language processing (NLP) task that involves identifying specific mentions in a given text and classifying them into predefined entity types, such as location, organisation, or person \cite{Lample2016_NER_Definition,Seow2025_review_of_NER}. NER is a well-studied problem in the literature, with early approaches relying on rules, dictionaries, and statistical taggers \cite{Nasar2021_rule_based_review_of_NER}, whereas modern approaches predominantly use deep-learning-based methods \cite{Li2022_DeepL_review_of_NER}. Recent progress is driven by contextual encoders, particularly Transformer-based models \cite{BERT_2019_Devlin,Contextualized_Embeddings_2020_BERT,Kaur2024_Bert-Ner}. Among these methods, span-based NER has become more prevalent because, unlike token-level sequence tagging, it identifies entities by scoring spans directly \cite{Zaratiana2022_span_based_better,Zhang2023_Neural_Span_Based}. Spans are often represented using features derived from the contextual embeddings of their boundary tokens (first and last) \cite{Su2022_span_representation_method}, typically produced by a pretrained language model (e.g., BERT \cite{BERT_2019_Devlin} or RoBERTa \cite{RoBERTa_2019_Liu}). This span-level modelling tends to improve robustness to out-of-vocabulary words \cite{Fu2021_SpanNER} and to missing or unlabelled entities arising from annotation errors \cite{li2021_Span_better_ICLR}.

The downstream task following NER in an ER pipeline is EL, which takes the detected entity mentions and maps each one to its corresponding entity in a KB, bridging unstructured language with structured knowledge for retrieval and reasoning \cite{Christophides2020_end_to_end_ER}. EL is commonly implemented as a two-stage process: (i) candidate generation (CG), also known as blocking, which retrieves a set of plausible KB entities using the mention while reducing the number of candidate descriptions to be compared, and (ii) entity matching, which ranks these candidates to resolve ambiguity using a set of functions (e.g., description similarity) \cite{Papadakis2020_ER_Blocking_Filtering}. Recent neural EL systems attempt to reduce reliance on hand-crafted similarity features by learning dense mention–context and entity representations (often using attention encoders) \cite{Sevgili2022_Neural_EL}. However, they typically remain challenged by long-tailed entity distributions (i.e., a few entities occur very frequently while many others are rare) and KB incompleteness (including NIL/unlinkable mentions), and can be computationally expensive, especially when using cross-encoder scoring for candidate ranking \cite{Shen2015_EL_with_KB,Wang2021_Knowledge_graph_quality,Sevgili2022_Neural_EL}.

Many ER systems still couple NER and EL through point decisions: EL CG is conditioned on the single best NER span and type, rather than a calibrated uncertainty signal over alternative spans/types \cite{kolitsas-etal-2018-end,tedeschi-etal-2021-named-entity}. This design is brittle because mention resolution is often underdetermined: small boundary differences change the mention string used for alias lookup and string matching during CG, and type errors can exclude otherwise plausible candidates when type constraints are applied \cite{tedeschi-etal-2021-named-entity,shen-wang-han-2015-entitylinking,hachey-etal-2013-evaluating}. The result is error propagation: early ambiguity becomes a hard commitment that downstream linking cannot easily recover from \cite{kolitsas-etal-2018-end}.

Uncertainty arises at multiple stages in the ER pipeline, and is therefore a meaningful signal, not just noise. At the NER stage, uncertainty typically appears as (i) boundary uncertainty (where the mention starts/ends) and (ii) label uncertainty (which entity type applies); recent works have explored uncertainty estimation and calibration for sequence labelling and NER \cite{hashimoto-etal-2024-data,He_2024_uncertainty_transmission,jagannatha-yu-2020-calibrating,Zhang2024_LinkNER_Uncertainty}. At the EL stage, uncertainty is driven by the knowledge base and data distribution itself: aliases are shared across entities, descriptions can be sparse, many entities are rare and therefore poorly represented, and some mentions are simply unlinkable (NIL) \cite{hoffart-etal-2011-robust,hachey-etal-2013-evaluating,shen-wang-han-2015-entitylinking}. In practice, CG involves balancing high recall against the introduction of irrelevant candidates through choices such as exact versus fuzzy matching, similarity thresholds, and pruning rules \cite{shen-wang-han-2015-entitylinking}. When upstream evidence is weak, however, a fixed CG strategy can either miss the gold entity or admit too many irrelevant candidates \cite{hachey-etal-2013-evaluating}. Uncertainty estimation for neural NLP has been widely studied, with Monte Carlo (MC) dropout providing a standard approximate Bayesian approach \cite{gal-ghahramani-2016-dropout}. For structured prediction, recent works target calibrated confidence and uncertainty scoring for outputs relevant to NER \cite{hashimoto-etal-2024-data,He_2024_uncertainty_transmission,jagannatha-yu-2020-calibrating}. However, comparatively little work uses NER uncertainty as an explicit control signal to adapt the EL CG policy, despite evidence that NER decisions materially affect downstream linking \cite{tedeschi-etal-2021-named-entity}.

Therefore, this work addresses that gap with an uncertainty-preserving design. We (i) extract mention-level uncertainty signals from a BERT-based NER model under MC dropout (boundary confidence and entropy) and (ii) leverage them to improve CG in EL. By preserving these uncertainty signals rather than reducing NER outputs to a single point estimate, the downstream EL stage is better able to support informed decision-making under uncertainty. For example, candidate-retrieval behaviour can be adapted when boundary confidence is low or entity-type entropy is high, thereby enabling more appropriate downstream decisions such as linking, abstaining from linking, or flagging the mention for further review instead of forcing an early commitment to an uncertain match. The contributions of this work are summarised as follows: \footnote{The experimental code and materials required to reproduce the results are publicly available at: \url{https://github.com/Nasser-Alkhulaifi/uncertainty-aware-cti-linking}.}

\begin{itemize}[leftmargin=15pt, itemsep=-2pt]
    \item Extraction of uncertainty signals at the NER stage by augmenting the model with two auxiliary span-boundary prediction heads (start/end), followed by MC dropout to estimate predictive uncertainty.
    \item Use of these proxies to adapt EL CG and improve gold-entity retention in a cyber threat intelligence setting (i.e., support better-informed downstream decisions under uncertainty).
\end{itemize}

The structure of the remaining sections of this paper is as follows: \autoref{Methods} describes how uncertainty signals are captured and leveraged, \autoref{Experimental Design} details the datasets, experimental setup, and evaluation metric, \autoref{Results and Discussion} analyses and discusses the results, and \autoref{Conclusion} summarises the key findings, limitations, and directions for future work.


\begin{figure*}[!t]
\centering
\includegraphics[width=\textwidth]{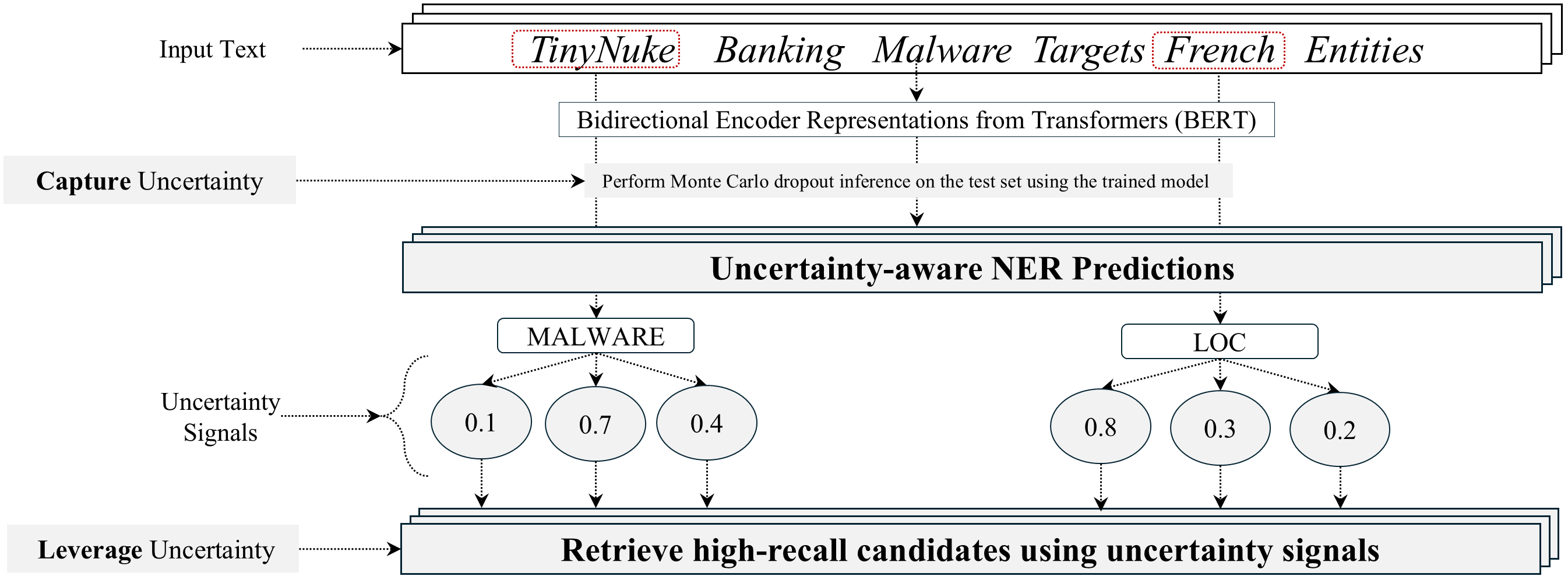}
\caption{Overview of our uncertainty-aware NER-to-EL approach: (i) capture and extract uncertainty signals in the NER stage and (ii) leverage these signals into the candidate retrieval stage to enable more robust and transparent linking decisions. The three captured uncertainty signals are (1) the MC-averaged start boundary probability, (2) the MC-averaged end boundary probability, and (3) the mention-level De Luca--Termini entropy over the normalised possibilistic span distribution. See \autoref{Methods} for more details on this approach and \autoref{Experimental Design} for the experimental procedure.}
\label{fig:our_approach}
\end{figure*}

\section{Methods}
\label{Methods}

This section describes how uncertainty signals are extracted and leveraged across the NER-to-EL components of the text-oriented ER workflow, as shown in \autoref{fig:our_approach}. \autoref{Capturing Uncertainty Signals} explains how uncertainty signals are captured at the NER level using a BERT-based approach augmented with MC dropout to estimate predictive uncertainty over entity spans and types. \autoref{Using Uncertainty Signals} explains how these signals are leveraged to retrieve candidates at the EL stage.


\subsection{Capturing Uncertainty Signals}
\label{Capturing Uncertainty Signals}

To capture uncertainty at the NER stage, a BERT-based model was used to produce contextual token representations and decode them with a Begin--Inside--Outside (BIO) tagging head \cite{Zhu2022_Boundary_NER}, plus two auxiliary boundary heads (start/end). This separates token type assignment from boundary localisation: the BIO head uses softmax because BIO labels are mutually exclusive, while the boundary heads use sigmoid because start/end are independent token-wise binary events (and can both be true for single-token mentions). BIO decoding remains the primary extraction mechanism; boundary heads provide additional boundary-confidence signals.

For uncertainty estimation, we apply MC dropout \cite{gal-ghahramani-2016-dropout,Shorinwa2025_LLM_MC_as_UQ} at inference time and run \(M\) stochastic passes. For BIO labels, we use the MC predictive distribution:

\begin{equation}
\label{eq:mc_predictive_distribution_bio}
\bar{p}_t(c)=\frac{1}{M}\sum_{m=1}^{M}\mathrm{softmax}(z_{t,m})_c
\end{equation}
where \(t\) indexes tokens, \(m\) indexes MC dropout passes, \(c\) is a BIO class, \(z_{t,m}\) are the BIO logits at token \(t\) in pass \(m\), and \(\mathrm{softmax}(z_{t,m})_c\) denotes the predicted probability of class \(c\) under that pass.

\begin{equation}
\label{eq:mc_boundary_start_mean}
\bar{q}^{start}_t=\frac{1}{M}\sum_{m=1}^{M} q^{start}_{t,m}
\end{equation}
where \(q^{start}_{t,m}\in[0,1]\) is the sigmoid output for the start-boundary head at token \(t\) in pass \(m\), and \(\bar{q}^{start}_t\) is its MC-averaged start-boundary probability.

\begin{equation}
\label{eq:mc_boundary_end_mean}
\bar{q}^{end}_t=\frac{1}{M}\sum_{m=1}^{M} q^{end}_{t,m}
\end{equation}
where \(q^{end}_{t,m}\in[0,1]\) is the sigmoid output for the end-boundary head at token \(t\) in pass \(m\), and \(\bar{q}^{end}_t\) is its MC-averaged end-boundary probability.

Hence, the three captured signals are: (1) MC-averaged start-boundary probability \(\bar{q}^{start}_t\), (2) MC-averaged end-boundary probability \(\bar{q}^{end}_t\), and (3) mention-level De Luca--Termini entropy\footnote{De Luca \& Termini entropy \cite{DELUCA1993_Entropy} on possibilistic distribution; measures average fuzziness of classification across span tokens; calculated after normalising probabilities to possibilities.}, computed from the normalised possibilistic form of \(\bar{p}_t(c)\) and averaged across tokens in the predicted mention span. Together, these signals capture complementary uncertainty about boundary reliability and entity-type ambiguity, and are passed to the downstream EL stage.

It is worth clarifying that, in this work, \textit{uncertainty} refers to the instability of the NER model's span and type predictions across repeated MC-dropout inference passes. By keeping dropout active at inference and running multiple stochastic forward passes, we approximate an ensemble of closely related models and observe how consistently these runs agree on BIO labels and on the start/end boundary estimates produced by the two auxiliary span-boundary prediction heads. Accordingly, the uncertainty considered here is upstream NER uncertainty about where a mention begins and ends and which entity type it should receive. It is not a direct confidence score for the final EL decision; for example, we do not estimate quantities such as \textit{this mention links to ATT\&CK entity \(e\) with probability \(p\)}. Nor do we propagate a full distribution over alternative spans or entity types into EL. Instead, we average the variability in the NER outputs using the three aforementioned mention-level proxy signals derived from the MC predictive BIO distribution over the predicted span tokens. These signals are used as heuristics for span reliability and type ambiguity: the boundary probabilities indicate confidence in the chosen start and end tokens, whereas the entropy score is a relative measure of type ambiguity, with higher values indicating greater ambiguity rather than a calibrated probability.


\subsection{Leveraging Uncertainty Signals}
\label{Using Uncertainty Signals}

After uncertainty proxies have been derived at the NER stage, the EL candidate generator receives each predicted mention together with its boundary-confidence, entropy, and predicted-type information.
For each mention, EL receives the \texttt{predicted entity type}, \texttt{start/end boundary probabilities}, and \texttt{mention entropy}; these signals are then used for CG, as explained in the following section.


\subsubsection{Candidate Generation Approaches}
\label{Candidate-generation approaches}

In this stage, different CG methods are used with the same high-level pipeline: start from exact alias matching, then expand candidates with fuzzy matching using \texttt{RapidFuzz.Ratio} \footnote{RapidFuzz is an open-source C++/Python library for efficient approximate string matching based on edit-distance and related similarity metrics. It computes normalised similarity scores between strings using optimised implementations of algorithms such as Levenshtein distance, enabling fuzzy matching. \url{https://github.com/rapidfuzz/RapidFuzz}}. The key difference between the methods is how the fuzzy cutoff is defined (fixed vs uncertainty-adaptive). This cutoff is the minimum similarity score required to retain a candidate alias, thereby pruning weak matches and reducing the number of comparisons that must be considered downstream \cite{Papadakis2020_ER_Blocking_Filtering}.

For mention $m$, let $b_m=(\bar q^{start}_{s_m}+\bar q^{end}_{e_m})/2$ and let $H_m$ be its mean token-level fuzzy entropy. The implemented cutoffs, on the 0--100 similarity scale, are
\[
\begin{aligned}
\tau_{FT}&=70,\\
\tau_{BA}(m)&=\max\{40,\lfloor70b_m\rfloor\},\\
\tau_{EA}(m)&=\max\{40,70-\lfloor30H_m\rfloor\}.
\end{aligned}
\]
Here $H_m$ sums $-[\pi\ln\pi+(1-\pi)\ln(1-\pi)]$ over BIO classes after $\pi_t(c)=\bar p_t(c)/\max_j\bar p_t(j)$, then averages over span tokens; it is not normalised to $[0,1]$, and $H_m\geq1$ reaches the cutoff floor. Each fuzzy search retains at most 50 aliases before entity-ID deduplication; the shared ranker returns at most 50 entities. DS uses the union of BA and EA candidates.The values 70 and 30 were chosen as heuristic design parameters rather than optimised hyperparameters: 70 establishes the baseline similarity cutoff and scales boundary-based adaptation, while 30 controls the degree of entropy-driven relaxation, allowing uncertainty to influence candidate retrieval without removing the minimum similarity constraint.

Generally speaking, a higher cutoff is stricter (fewer candidates); a lower cutoff is looser (more candidates). It is worth emphasising that in this stage, the aim is to achieve high recall, rather than high precision (i.e., the objective of CG is to maximise coverage of plausible entities so that the gold entity is not discarded before the ranking stage; consequently, achieving high recall is essential \cite{Efthymiou2015_blocking,Borchert2024_high_recall_is_essential}). In other words, recall is prioritised here because CG is a blocking step: if the gold entity is filtered out at this stage, no downstream ranker can recover it. Note that, while lowering the fuzzy cutoff enlarges the candidate set, which is essential for recall-oriented candidate generation \cite{Borchert2024_high_recall_is_essential,Fang2020_high_recall_is_essential}, it may also increase downstream computational cost and make final linking less selective by introducing more entities. Accordingly, the present study treats recall as the primary evaluation criterion for CG specifically. Precision remains important for end-to-end EL, but it depends jointly on CG and ranking quality. For this reason, the ranking stage, described in \autoref{Ranking Strategy}, is intentionally held fixed across methods so that performance differences can be attributed specifically to uncertainty-aware retrieval. Under this design, the central question can be simplified to whether uncertainty-aware CG preserves the correct entity more reliably before ranking, rather than prematurely eliminating it (i.e., the goal is to ensure the correct entity survives into the ranking stage).

We used five different CG methods to form a controlled progression from uncertainty-discarding to uncertainty-preserving retrieval: (i) Exact Linking (EX) and (ii) Fixed-Threshold Fuzzy Linking (FT) as baselines, then (iii) Boundary-Aware Fuzzy Linking (BA), (iv) Entropy-Aware Fuzzy Linking (EA), and (v) Dual-Signal Fuzzy Linking (DS) as uncertainty-aware variants. This design enables a principled comparison of retrieval policies by progressively increasing the amount of uncertainty information used at CG: from no uncertainty (EX, FT), to a single uncertainty cue (BA, EA), to both cues jointly (DS). This progression examines not only whether uncertainty improves retrieval, but also which cue contributes most and whether their combination provides complementary gains. These methods are summarised in \autoref{tab:candidate_generation_comparison} and detailed as follows:

\begin{table*}[!t]
\centering
\scriptsize
\caption{Summary of candidate generation approaches. See \autoref{Candidate-generation approaches} for more details.}
\label{tab:candidate_generation_comparison}
\renewcommand{\arraystretch}{1}
\begin{tabular}{>{\raggedright\arraybackslash}p{0.05\linewidth}|
                >{\centering\arraybackslash}p{0.05\linewidth}|
                >{\centering\arraybackslash}p{0.05\linewidth}|
                >{\centering\arraybackslash}p{0.05\linewidth}|
                >{\centering\arraybackslash}p{0.12\linewidth}|
                >{\raggedright\arraybackslash}p{0.21\linewidth}|
                >{\raggedright\arraybackslash}p{0.23\linewidth}}
\toprule
\textbf{Method}
& \textbf{Exact}
& \textbf{Fuzzy}
& \textbf{Cutoff Type}
& \textbf{Uncertainty Signal}
& \textbf{Adaptive Behaviour}
& \textbf{Final Candidate Set} \\
\midrule
EX & \cmark & \xmark & None & None & No adaptation & Exact alias matches only \\
FT & \cmark & \cmark & Fixed & None & Fixed cutoff for all mentions & Exact + fuzzy matches above fixed threshold \\
BA & \cmark & \cmark & Adaptive & Start/end boundary probabilities & Lower cutoff when start/end boundary confidence is low & Exact + fuzzy matches with boundary-adjusted threshold \\
EA & \cmark & \cmark & Adaptive & Mention entropy & Lower cutoff when entropy is high & Exact + fuzzy matches with entropy-adjusted threshold \\
DS & \cmark & \cmark & Adaptive & Boundary + entropy & Combine boundary- and entropy-based retrieval cues & Merge the boundary- and entropy-based candidate sets, then remove duplicates \\
\bottomrule
\end{tabular}
\end{table*}



\begin{enumerate}[label=(\Roman*), leftmargin=*, itemsep=0.2em]
  \item \textbf{Exact Linking Baseline}
  \begin{itemize}
    \item \textit{Input signals used:} mention text only.
    \item \textit{Retrieval rule:} retrieve candidates only if the mention exactly matches a KB alias.
    \item \textit{Rationale:} uncertainty-discarding baseline with high precision but limited coverage.
    \item \textit{Expected effect:} lowest recall when lexical variation exists.
  \end{itemize}

  \item \textbf{Fixed-Threshold Fuzzy Linking}
  \begin{itemize}
    \item \textit{Input signals used:} mention text only.
    \item \textit{Retrieval rule:} exact alias candidates + fuzzy alias candidates above fixed cutoff.
    \item \textit{Rationale:} isolates the benefit of fuzzy lexical matching without uncertainty adaptation.
    \item \textit{Expected effect:} higher recall than EX baseline, but same retrieval strictness (i.e., cutoff) for both clear and ambiguous mentions.
  \end{itemize}

  \item \textbf{Boundary-Aware Fuzzy Linking}
  \begin{itemize}
    \item \textit{Input signals used:} mention text and start/end boundary probabilities.
    \item \textit{Retrieval rule:} exact alias candidates + fuzzy candidates with a mention-specific cutoff adapted from start/end boundary probabilities, treated jointly as a boundary-confidence signal.
    \item \textit{Rationale:} when start/end boundary confidence is low, the mention string is less reliable, so retrieval is broadened by lowering the cutoff.
    \item \textit{Expected effect:} improved recall for boundary-uncertain mentions.
  \end{itemize}

  \item \textbf{Entropy-Aware Fuzzy Linking}
  \begin{itemize}
    \item \textit{Input signals used:} mention text and entropy.
    \item \textit{Retrieval rule:} exact alias candidates + fuzzy candidates with a mention-specific cutoff adapted from entropy.
    \item \textit{Rationale:} when mention-level ambiguity is high, retrieval is broadened by lowering the cutoff.
    \item \textit{Expected effect:} improved recall for semantically ambiguous mentions.
  \end{itemize}

  \item \textbf{Dual-Signal Fuzzy Linking}
  \begin{itemize}
    \item \textit{Input signals used:} mention text, start/end boundary probabilities, and entropy.
    \item \textit{Retrieval rule:} build one candidate set with Boundary-Aware and one with Entropy-Aware, then merge and deduplicate.
    \item \textit{Rationale:} combines complementary uncertainty cues (span-boundary reliability and label ambiguity).
    \item \textit{Expected effect:} broadest uncertainty-aware coverage among adaptive methods.
  \end{itemize}
\end{enumerate}




\subsubsection{Ranking Strategy}
\label{Ranking Strategy}
All methods used for CG apply the same ranking (scoring) strategy. For each generated candidate entity, the ranker: (i) computes the maximum string similarity between the mention text and all aliases of that entity (from the candidate set), (ii) applies a type-consistency weight from the predicted entity type, and (iii) sorts candidates by the final score in descending order (i.e., $\text{string\_similarity} \times \text{type\_weight}$). Therefore, performance differences are attributable to CG approaches (exact, fixed fuzzy, or uncertainty-adaptive), not to ranking changes. The ranking strategy is computed as:

{\small
\begin{equation}
\label{eg:ranking}
\begin{aligned}
\mathrm{score}(e \mid m)
&= \left(
\max_{a \in \mathrm{Aliases}(e)}
\frac{\mathrm{RapidFuzz.Ratio}(m,a)}{100}
\right) \\
&\quad \times w_{\mathrm{type}}(e)
\end{aligned}
\end{equation}
}

where \(m\) is the mention text, \(e\) is a candidate entity, \(a\) is an alias of \(e\), and \(w_{\mathrm{type}}(e)\) is the type-consistency weight applied to candidate \(e\):

{\small
\[
w_{\mathrm{type}}(e)=
\begin{cases}
1.0, & \begin{aligned}[t]
\text{if KB type of } e \text{ matches}\\
\text{predicted mention type}
\end{aligned}\\
0.5, & \text{otherwise.}
\end{cases}
\]
}

\textbf{Steps:}
\begin{itemize}
    \item For each candidate entity $e$, consider all its aliases.
    \item For every alias $a \in \mathrm{Aliases}(e)$, compute the normalised string similarity $\mathrm{RapidFuzz.Ratio}(m,a)/100$ between the mention $m$ and alias $a$.
    \item Select the maximum similarity score across all aliases.
    \item Multiply this maximum score by the type-consistency weight $w_{\mathrm{type}}(e)$.
\end{itemize}

Notably, we \emph{deliberately} do not use uncertainty signals as a direct ranking factor, because doing so is not methodologically appropriate in our current setup. The uncertainty signals extracted from the NER stage are mention-level (i.e., one value per mention, shared across all candidate entities for that mention). Injecting a shared mention-level value into ranking would affect all candidate scores equally, leaving their relative order unchanged and therefore not improving rank discrimination. For this reason, by design, we apply uncertainty at the CG stage, where it can meaningfully change which candidates enter the ranking stage. A worked example with four illustrative scoring scenarios is provided in Table~\ref{tab:ranking_cases} to show how similarity and type-consistency jointly determine the final rank score.


\begin{table*}[!t]
\centering
\scriptsize
\caption{Illustrative ranking cases for a single mention (\texttt{TinyNuke}, predicted type \texttt{MALWARE}). Cases 1 and 3 isolate the effect of lexical similarity when the KB type matches the predicted type (both \texttt{MALWARE}, but with different \texttt{Fuzz.Ratio} values). Cases 1 and 2 isolate the effect of the type-consistency penalty under identical lexical matching, since in both cases the mention exactly matches the candidate alias \texttt{TinyNuke}, but the KB type differs. Case 4 combines both a weaker lexical match and a type mismatch, yielding the lowest final score.}
\label{tab:ranking_cases}
\renewcommand{\arraystretch}{1}
\begin{tabular}{>{\centering\arraybackslash}p{0.04\linewidth}|
                >{\raggedright\arraybackslash}p{0.35\linewidth}|
                >{\centering\arraybackslash}p{0.1\linewidth}|
                >{\centering\arraybackslash}p{0.1\linewidth}|
                >{\centering\arraybackslash}p{0.1\linewidth}|
                >{\centering\arraybackslash}p{0.1\linewidth}}
\toprule
\textbf{Case} & \textbf{Candidate Alias Example} & \textbf{KB Type} & \(\mathrm{Fuzz.Ratio}\) & \(w_{\mathrm{type}}\) & \textbf{Final Score} \\
\midrule
1 & \texttt{TinyNuke} (exact alias, type match) & MALWARE & 1.00 & 1.0 & 1.00 \\
2 & \texttt{TinyNuke} (exact alias, type mismatch) & TOOL & 1.00 & 0.5 & 0.50 \\
3 & \texttt{MiniDuke} (lexically similar, type match) & MALWARE & 0.625 & 1.0 & 0.625 \\
4 & \texttt{CosmicDuke} (some overlap, type mismatch) & TOOL & 0.64 & 0.5 & 0.32 \\
\bottomrule
\end{tabular}
\end{table*}



\section{Experimental Design}
\label{Experimental Design}

This section describes the experimental design used in this work. It presents the datasets in \autoref{datasets}, the experimental setup in \autoref{Experimental Setup}, and the evaluation metric in \autoref{Evaluation Metric}.



\subsection{Datasets}
\label{datasets}

This study uses the AnnoCTR dataset \cite{annoctr2024}, which contains 400 cyber threat intelligence reports annotated by domain experts. The annotations cover both general named entities (e.g., organisations, locations) and cybersecurity-specific entities (e.g., malware, tools, techniques), including implicit mentions of MITRE ATT\&CK concepts \cite{MITRE_ATTCK_Design_Philosophy_2020}. 

For the NER stage (capturing uncertainty), the annotated token/span labels in AnnoCTR are used to train and evaluate mention boundary detection and entity-type prediction, from which uncertainty signals (e.g., boundary confidence and entropy) are derived. For the EL stage (leveraging uncertainty), the resulting predicted mentions and uncertainty signals are used to retrieve candidates against the MITRE ATT\&CK knowledge base, enabling end-to-end evaluation of span detection, type prediction, and EL.


\subsection{Experimental Setup}
\label{Experimental Setup}

For the NER stage, we use the official AnnoCTR split \cite{annoctr2024}, obtained from the Hugging Face datasets library\footnote{Hugging Face is a widely used machine-learning platform and software ecosystem for hosting and accessing models, datasets, and related tools; see \url{https://huggingface.co/docs/datasets/index}. In this work, specifically, the \texttt{datasets} library is used to retrieve the published dataset split.} (7,570 train, 1,550 validation, 3,059 test). We use three BERT-based models (BERT-cased, BERT-uncased \cite{BERT_2019_Devlin}, RoBERTa \cite{RoBERTa_2019_Liu} for the NER task. The models use an encoder with three heads: BIO token classification, start-boundary prediction, and end-boundary prediction. Training uses a multi-task loss (token cross-entropy + boundary binary cross-entropy) with boundary weight \(\lambda=0.2\), chosen as a conservative setting to inject boundary supervision without overpowering the main BIO objective. We use a learning rate \(1\times10^{-5}\), batch size 16, a maximum of 100 epochs, and early stopping patience 20; the best checkpoint is selected based on the F1 score on the validation set. We did not run an exhaustive grid or random hyperparameter search; this is intentional because the study objective is to reliably capture uncertainty signals for downstream EL, rather than fully optimise peak NER accuracy. At inference time, uncertainty is estimated with MC dropout (\(M=20\) stochastic passes), consistent with settings used in related uncertainty estimation literature \cite{hashimoto-etal-2024-data,Liu2023_MC_Dropout_20}. The resulting MC-averaged boundary probabilities are then leveraged as uncertainty signals to the EL stage. Experiments were implemented in Python using PyTorch and the Hugging Face Transformers/Datasets libraries.


\subsection{Evaluation Metric}
\label{Evaluation Metric}

For evaluation, we use mention-level \(\mathrm{Recall@}K\) with \(K\in\{1,10,50\}\), following standard EL retrieval evaluation practice \cite{Shen2015_EL_with_KB,Kartchner2023_EL_recall,monath_2023_google_paper_recall}:

\begin{equation}
\label{eq:recall_at_k}
\mathrm{Recall@}K
=
\frac{\#\{\text{correct mentions at }K\}}
{\#\{\text{evaluable mentions}\}}
\end{equation}

In \autoref{eq:recall_at_k}, a mention is counted as correct at \(K\) if at least one gold ATT\&CK entity ID for that mention appears among its top-\(K\) predicted candidates. Here, evaluable mentions are those for which a gold ATT\&CK entity ID is available and for which the method produces a prediction. We report \(\mathrm{Recall@1}\), \(\mathrm{Recall@10}\), and \(\mathrm{Recall@50}\), which indicate whether at least one gold ATT\&CK entity ID is retrieved within the top 1, top 10, and top 50 candidates, respectively. This metric provides empirical evidence for our core hypothesis by quantifying whether uncertainty-preserving retrieval retains the correct entity in the candidate set more reliably than uncertainty-discarding baselines.

\section{Results and Discussion}
\label{Results and Discussion}

This section presents an analysis and discussion of the results. \autoref{Named Entity Recognition} discusses the overall performance of the NER task, including performance by model and by label. Subsequently, \autoref{Entity Linking} analyses the EL results, showing how uncertainty captured in the upstream task (i.e., the NER stage) contributes to high recall in the downstream task (i.e., the EL stage).

\begin{figure*}[!t]
\centering
\includegraphics[width=1\linewidth]{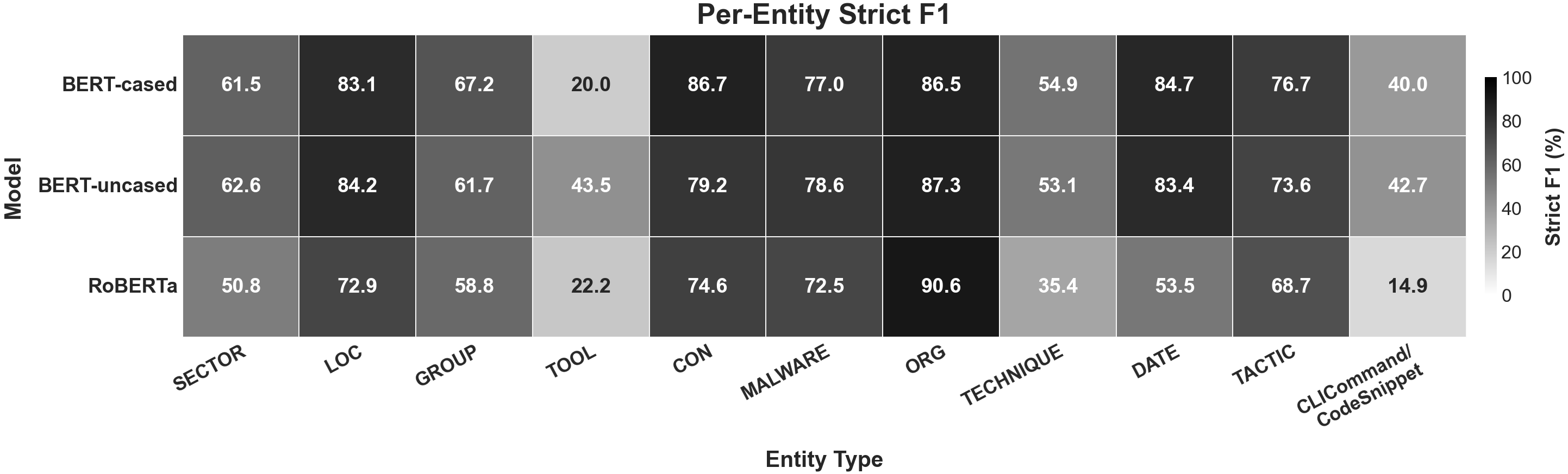}
\caption{Per-entity strict F1 scores for the three NER models, shown as a grayscale heatmap with models on the x-axis and entity types on the y-axis. Darker cells indicate higher strict F1 under the strict Nervaluate schema \cite{nervaluate}, where a prediction is counted as correct only when both the entity span boundaries and the entity type exactly match the ground truth; partial overlaps are treated as incorrect.}
\label{fig:nervaluate_transposed}
\end{figure*}

\subsection{Named Entity Recognition}
\label{Named Entity Recognition}

As shown in \autoref{fig:nervaluate_transposed}, NER performance varies substantially by entity type, indicating that errors are concentrated in particular semantic categories rather than being distributed uniformly across the label space. In particular, entity types such as \texttt{LOC}, \texttt{CON}, \texttt{ORG}, and \texttt{DATE} achieve comparatively high strict F1, suggesting that these mentions are generally easier to detect with the correct boundaries and assign the correct type, whereas \texttt{TOOL}, \texttt{TECHNIQUE}, and \texttt{CLICommand/CodeSnippet} remain much more challenging to identify. \autoref{fig:nervaluate_transposed} also shows model-specific variation within these labels: \texttt{TOOL}, for example, differs substantially between BERT-cased and BERT-uncased (20.0\% vs. 43.5\%), suggesting that casing and tokeniser sensitivity may interact with the surface form of cyber threat terms. For production deployment, this motivates fine-tuning and validating the NER model on target CTI reports, including their casing conventions, rather than assuming that a single pretrained encoder configuration will transfer uniformly across entity types. This is important in the context of the present work because these harder-to-predict categories are also those for which incorrect NER outputs are more likely to propagate downstream into EL through boundary or type errors, thereby reducing candidate coverage under conventional point-estimate pipelines. Consequently, the figure reinforces the motivation for uncertainty-preserving ER; the challenge is not simply average NER quality, but the existence of entity-type-specific failure modes that require downstream CG to respond adaptively when upstream predictions are less reliable.


\begin{figure*}[t]
\centering
\includegraphics[width=1\linewidth]{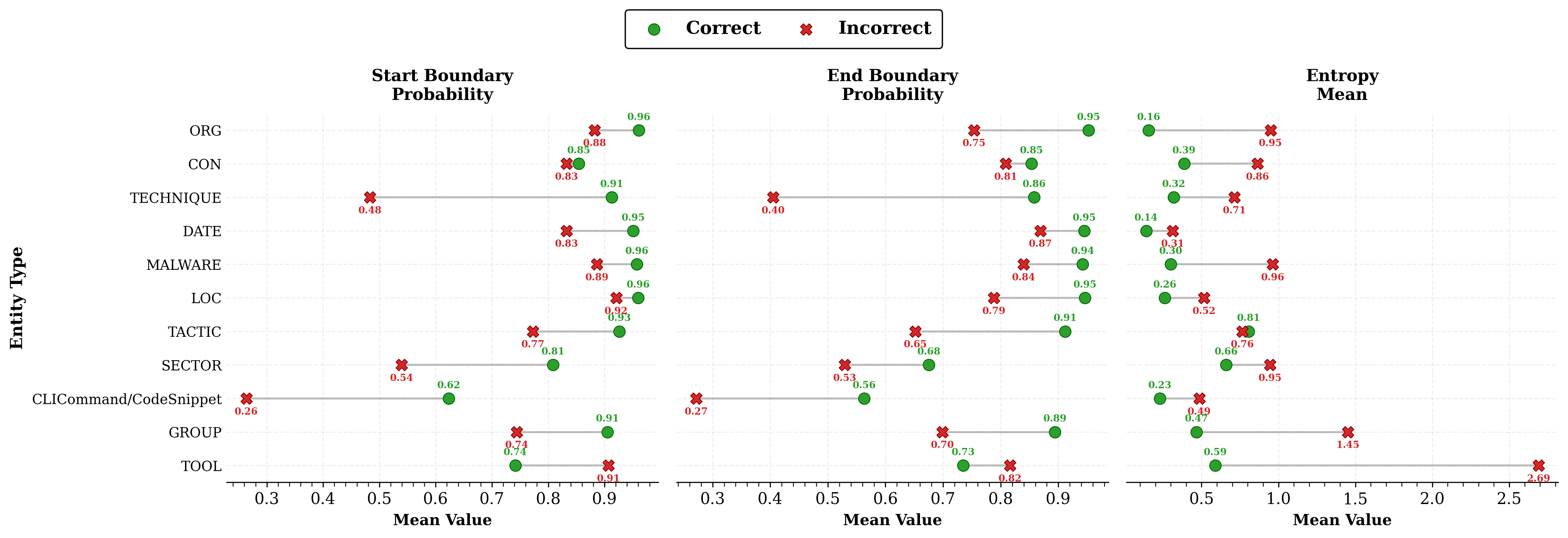}
\caption{Boundary confidence and entropy by entity type (correct vs. incorrect predictions). Each row represents an entity type. The three panels show the mean start/end boundary probability and mean entropy. Green circles indicate correct predictions and red crosses indicate incorrect ones, connected by a grey line to emphasise the gap. Larger horizontal gaps reveal entity types where the model's confidence drops for incorrect spans, while the entropy panel highlights which types exhibit the greatest uncertainty when predictions fail. Together, the three panels identify which entity types are most amenable to confidence-based filtering and where boundary detection remains challenging.}

\label{fig:boundary_entropy_dumbbell}
\end{figure*}

\autoref{fig:boundary_entropy_dumbbell} highlights that, across most entity types, incorrect NER predictions are characterised by a consistent shift toward lower start/end boundary probabilities and higher mean entropy, indicating that model failures are typically associated with both weaker span localisation and greater label ambiguity. This separation is especially pronounced for types such as \texttt{TECHNIQUE}, and \texttt{CLICommand/CodeSnippet}, where incorrect predictions exhibit markedly lower boundary confidence than correct ones, suggesting that these categories are particularly sensitive to boundary uncertainty and are therefore strong candidates for uncertainty-aware downstream handling. At the same time, the entropy panel reveals an important complementary pattern: most notably for \texttt{TOOL}, incorrect predictions show a very large increase in entropy even when boundary probabilities remain relatively high, implying that these errors are driven less by span detection failure than by type-level ambiguity. This observation, while dataset-specific, supports the central design choice of preserving multiple uncertainty signals from NER rather than collapsing predictions to a single point estimate, because boundary confidence and entropy capture different failure modes and therefore provide distinct evidence for adapting CG in the EL stage.


\begin{figure*}
\centering
\includegraphics[width=1\linewidth]{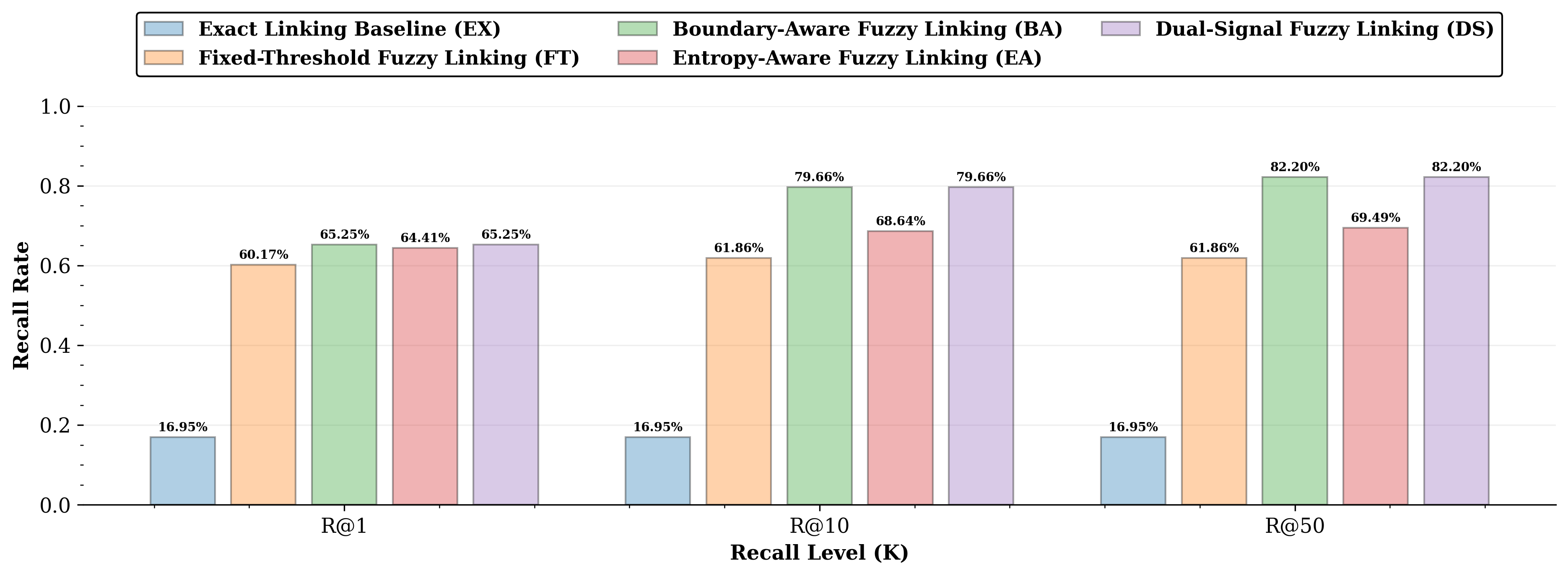}
\caption{Recall across different EL methods indicates whether the gold entity is retrieved within the top 1, top 10, and top 50 candidates, respectively. This provides empirical support for the study’s hypothesis by assessing whether uncertainty-preserving retrieval retains the correct entity in the candidate set more reliably than uncertainty-discarding baselines.}
\label{fig:Recall}
\end{figure*}

\subsection{Entity Linking}
\label{Entity Linking}

As shown in \autoref{fig:Recall}, uncertainty-aware CG considerably improves retrieval over uncertainty-discarding baselines. For example, compared with EX (R@10 = 16.95\%) and FT (R@10 = 61.86\%), boundary-aware retrieval reaches R@10 = 79.66\% and R@50 = 82.20\%, confirming that uncertainty-guided cutoff adaptation improves high-recall candidate coverage. Entropy-aware retrieval also improves over FT, but more modestly (R@10 = 68.64\%, R@50 = 69.49\%), suggesting that boundary confidence is the stronger retrieval control signal in this dataset. Notably, DS (boundary+entropy union) is numerically identical to BA across R@1, R@10, and R@50 (65.25\%, 79.66\%, and 82.20\%, respectively), indicating that, under the current signal calibration and thresholds, entropy-driven additions provide limited incremental benefit beyond boundary-adaptive retrieval. Overall, while these results are dataset-dependent, they support the core hypothesis of this work: preserving uncertainty signals from the upstream NER stage can improve EL in ER pipelines, and such signals are a useful component of the pipeline rather than an obstacle to be eliminated.

\begin{table*}[!t]
\centering
\scriptsize
\caption{Candidate generation and top-10 ranking results for different methods for the mention \texttt{TinyNuke}, selected as one mention-level example from the conducted experiment. See \autoref{Candidate-generation approaches} for details of candidate generation in each method, and \autoref{Ranking Strategy} for the calculation of each candidate's ranking score.}
\label{tab:candidate_cases}
\renewcommand{\arraystretch}{1}
\begin{tabular}{>{\centering\arraybackslash}p{0.05\linewidth}|>{\centering\arraybackslash}p{0.13\linewidth}|>{\raggedright\arraybackslash}p{.75\linewidth}}
\toprule
\textbf{Method} & \textbf{Candidates Found} & \textbf{Candidate Ranking (Top 10)} \\
\midrule
EX & 0 & No candidates \\
FT & 0 & No candidates \\
BA & 8 & 1. MiniDuke (0.3125) ; 2. TinyTurla (0.2941) ; 3. CosmicDuke (0.2941) ; 4. PinchDuke (0.2941) ; 5. OnionDuke (0.2941) ; 6. QUIETCANARY (0.2857) ; 7. GeminiDuke (0.2778) ; 8. TINYTYPHON (0.2778) \\
EA & 49 & 1. Winexe (0.4286) ; 2. MiniDuke (0.3125) ; 3. OnionDuke (0.2941) ; 4. PinchDuke (0.2941) ; 5. CosmicDuke (0.2941) ; 6. TinyTurla (0.2941) ; 7. QUIETCANARY (0.2857) ; 8. GeminiDuke (0.2778) ; 9. TINYTYPHON (0.2778) ; 10. HAMMERTOSS (0.2667) \\
DS & 49 & 1. Winexe (0.4286) ; 2. MiniDuke (0.3125) ; 3. TinyTurla (0.2941) ; 4. CosmicDuke (0.2941) ; 5. PinchDuke (0.2941) ; 6. OnionDuke (0.2941) ; 7. QUIETCANARY (0.2857) ; 8. GeminiDuke (0.2778) ; 9. TINYTYPHON (0.2778) ; 10. HAMMERTOSS (0.2667) \\
\bottomrule
\end{tabular}
\end{table*}

\autoref{tab:candidate_cases} provides an example of candidate retrieval for the mention \texttt{TinyNuke}. Uncertainty-discarding baselines (EX and FT) return no candidates, whereas uncertainty-aware retrieval returns non-empty candidate sets, with BA yielding a compact set (8 candidates) and EA and DS expanding to 49 candidates. This pattern indicates that entropy-adaptive cutoff relaxation is much more permissive than boundary-adaptive relaxation for this particular mention, and that DS is effectively dominated by the EA method here (its top-10 is nearly identical to EA), which is consistent with DS being a union of boundary- and entropy-derived candidate sets. Importantly, the top-ranked candidate under EA and DS is \texttt{Winexe} rather than \texttt{MiniDuke}, illustrating the ranking strategy explained in \autoref{Ranking Strategy} where the final score reflects both lexical similarity and type consistency, not lexical similarity alone. This case illustrates how uncertainty-adaptive retrieval changes candidate-set breadth and avoids an empty retrieval result; unless the gold ATT\&CK entity is explicitly marked in the table, the example should not itself be interpreted as evidence of successful gold-entity retrieval.

These recall gains, however, involve a selectivity trade-off. EX, FT, BA, EA, and DS retrieval return an average of 0.26, 9.02, 44.96, 26.5, and 44.96 candidates per evaluable mention, respectively (computed as the mean number of candidate entities retrieved from the KB for each evaluable mention under each method). This expansion is expected and desirable \cite{Efthymiou2015_blocking,Borchert2024_high_recall_is_essential,Fang2020_high_recall_is_essential}, within the present investigation, because CG is being evaluated as a coverage-preserving blocking step. Nonetheless, larger candidate sets increase ranking burden, require more comparisons, and may add irrelevant alternatives, raising the risk of erroneous final links when the system must choose one entity. In large-scale settings, large KBs and high mention volumes can make candidate-set increases accumulate into substantial runtime, memory, and inference costs, especially with a more expensive downstream ranker than the simple scorer used here. Accordingly, these results should be interpreted as improved CG rather than as evidence that larger candidate sets are universally preferable for end-to-end EL.

\section{Conclusion, Limitations and Future Work}
\label{Conclusion}

This work demonstrates that uncertainty in ER pipelines should not be treated as noise to be eliminated through early hard decisions, but rather as meaningful information that should be preserved and exploited throughout the pipeline. Conventional ER pipelines tend to represent the upstream task (i.e., NER) outputs as a single span and type prediction, leading downstream EL components to operate under an artificial assumption of certainty. As a result, valuable signals related to span ambiguity and type confusion are discarded, which can make ER systems brittle when confronted with uncertain real-world data.

To address this limitation, this paper presented a proof-of-concept uncertainty-aware EL approach in which uncertainty is treated as an inherent property of the data and explicitly propagated across pipeline stages. A BERT-based model with MC dropout was used to estimate predictive uncertainty over entity spans and types at the NER stage, and these signals were subsequently incorporated into CG during the EL stage.

Experimental results show that uncertainty-preserving retrieval produces higher-recall candidate sets than uncertainty-discarding baselines. In particular, recall improved from \(R@10 = 0.6186\) to \(0.7966\). These findings provide empirical evidence that preserving upstream ambiguity can improve downstream robustness in ER pipelines. The primary contribution of this work, therefore, lies not in presenting a fully optimised end-to-end ER system, but in establishing uncertainty preservation as a viable and beneficial design principle for future ER pipelines.

Despite the observed empirical gains in CG, several limitations should be noted. First, the experiments are conducted on a single domain-specific dataset (AnnoCTR \cite{annoctr2024}), so the findings may not generalise to domains with different entity distributions or KBs. Future work should evaluate the proposed approach on datasets beyond cyber threat intelligence to assess the cross-domain generalisability of the observed CG gains. Second, uncertainty is estimated only through MC dropout. Although this is a well-established approach \cite{Shorinwa2025_LLM_MC_as_UQ}, it is only one approximation and may not yield optimally calibrated uncertainty. Third, while a validation set was used for checkpoint selection at the NER stage, the study did not include exhaustive hyperparameter optimisation, since the objective was to assess the downstream value of uncertainty signals rather than maximise peak NER performance. 

Fourth, uncertainty is used only to adapt downstream CG and is not fed back to improve the upstream NER model itself; future work could investigate uncertainty-aware training or calibration \cite{He_2024_uncertainty_transmission}. Fifth, the EL component focuses specifically on CG under a fixed and relatively simple ranking strategy, so the observed gains should be interpreted as improvements in candidate coverage rather than full end-to-end linking performance. Relatedly, the adaptive cutoff design was not subjected to parameter-sensitivity analysis, and the current approach to combining boundary and entropy signals may not be optimal, as reflected in the limited additional benefit of the dual-signal variant (see \autoref{fig:Recall}). Finally, evaluation is centred on mention-level Recall@K because high recall is essential at the CG stage \cite{Efthymiou2015_blocking,Borchert2024_high_recall_is_essential,Fang2020_high_recall_is_essential}, where the principal failure mode is the premature discarding of the gold entity; that is, the goal is to ensure that the correct entity survives into the ranking stage. This does not necessarily imply that uncertainty-aware retrieval improves end-to-end precision; rather, it improves CG. Additionally, increasing the candidate set size can raise computational cost \cite{Papadakis2020_ER_Blocking_Filtering}; hence, a trade-off between maximising gold-entity retention at the CG stage and maintaining computational efficiency should be considered in future work.


\section*{\uppercase{Acknowledgements}}

This work was supported by the Turing's Defence and Security programme
through a partnership with the UK government in accordance with the framework agreement between GCHQ and The Alan Turing Institute.


\bibliographystyle{apalike}

{\small
\bibliography{references}
}

\end{document}